\pdfoutput=1

\documentclass[11pt]{article}

\usepackage{acl}

\usepackage{times}
\usepackage{adjustbox}
\usepackage{latexsym}
\usepackage{amssymb}
\usepackage[T1]{fontenc}

\usepackage[utf8]{inputenc}

\usepackage{microtype}
\usepackage{multirow}

\title{Exploring Answer Information Methods for Question Generation with Transformers}

\author{Talha Chafekar\\
  K.J. Somaiya College of Engineering\\
  \texttt{talha.c@somaiya.edu} \\\And
  Aafiya Hussain \\
  K.J. Somaiya College of Engineering\\
  \texttt{aafiya.h@somaiya.edu} \\\AND 
  Grishma Sharma \\
  K.J. Somaiya College of Engineering\\
  \texttt{grishma.sharma@somaiya.edu} \\\And 
  Deepak Sharma \\
  K.J. Somaiya College of Engineering\\
  \texttt{deepaksharma@somaiya.edu} \\
  }

\begin{document}
\maketitle
\begin{abstract}
There has been a lot of work in question generation where different methods to provide target answers as input, have been employed. This experimentation has been mostly carried out for RNN based models. We use three different methods and their combinations for incorporating answer information and explore their effect on several automatic evaluation metrics. The methods that are used are answer prompting, using a custom product method using answer embeddings and encoder outputs, choosing sentences from the input paragraph that have answer related information, and using a separate cross-attention attention block in the decoder which attends to the answer. We observe that answer prompting without any additional modes obtains the best scores across rouge, meteor scores. Additionally, we use a custom metric to calculate how many of the generated questions have the same answer, as the answer which is used to generate them.

\end{abstract}

\section{Introduction}
Question Generation involves generating questions given a passage with or without an answer prompt. It has many widespread applications like generating assessments \citep{ghanem2022question}, evaluating the reader’s understanding of the text by answering these questions, and intelligent tutoring systems \citep{shah2017automatic}.  Question generation is also used to aid question answering systems. \citet{duan2017question} and \citet{tang2017question} showed that question generation and question answering are dual tasks. The task of question generation consists of input from different modalities involving images, text, knowledge graphs, structured data. In text based question generation, various types of questions such as multiple choice, fill in the blanks, factoid, true and false questions, Wh questions, and questions requiring deep understanding of the text can be generated.

Fill in the blanks and multiple choice questions involve selecting keyphrases which can serve as answers. Multiple-choice questions, additionally require the generation of distractors. However, factoid and non-factoid questions involve natural language generation, thus requiring a deeper understanding of the source material. In scenarios where questions are generated from a passage, there is a one to many correspondence between the input passage and the generated question. This is because many questions can be generated depending on the context of the passage. This central idea can either be provided as an auxiliary input or be a part of the process which generates questions. This leads to two types of question generation methods, answer based question generation and answer agnostic question generation. In this paper, we analyze how incorporating different answer information methods affect the quality of questions generated using transformer architectures. 

We use transformers, specifically BART \citep{lewis2019bart}, as our base architecture for the task of question generation on the SQUAD dataset \citep{rajpurkar2016squad}. For our analysis, we carry out experiments with multiple ways to incorporate answer information, along with data manipulation of input text.

\section{Related Work}

The task of generating questions from text involves the use of sequence models like LSTM \citep{hochreiter1997long}, GRU \citep{cho2014learning}, RNN \citep{mcculloch1943logical} and transformer models like BART \citep{lewis2019bart}, Pegasus \citep{zhang2020pegasus}, T5 \citep{xue2020mt5}, etc. Previously, questions were created with the help of hand-crafted features from the text. \citet{du2017learning} used an LSTM for the task of question generation on SQUAD dataset \citep{rajpurkar2016squad}, which is an end to end trainable method. An attention mechanism is used so as to get a better representation of the outputs from the encoder LSTM. However, there is no answer information included separately as the input text is the sentences where the answer is located. \citet{yuan2017machine} follow a similar encoder-decoder approach using LSTMs, except they include a binary vector indicating if the word belongs to the answer or not. Linguistic features are also used \citep{harrison2018neural} in addition to word embeddings to improve the representation of text for the task. \citet{liu2019learning} make use of a dependency tree structure of the question to eliminate parts of the text which are not important for generating the question. \citet{gupta2019improving} use world knowledge using entity linking and fine grained entities in the form of concatenated vectors as an input to the encoder.

A position based approach for encoding answers has been used by \citet{sun2018answer}. which helps in generating words which are closer to the answer instead of generating words far away from the answer. A similar approach of using answer positioning is used by \citet{ma2020improving}. in their approach, except, they also use a sentence level semantic matching for encoder. Masking the answer with a special token as proposed by \citet{kim2019improving} makes the model generate words only relevant to the question and not to the answer, as some generation outputs consist of words from the answer. Content selection methods \citep{du2017identifying, subramanian2017neural, wang2019weak, cho2019mixture} identify content which is required for generating questions. Transformer based models were also used for end to end question generation where \citet{lopez2020transformer} use a GPT-2 decoder only model for generating questions from SQUAD dataset.

\section{Problem Statement}
Given an input passage \begin{math} \mathbf{X = x_1,...,x_n},  \end{math} we aim to generate a question \begin{math} \mathbf{Y^{'} = y^{'}_1,...,y^{'}_m} \end{math} where \begin{math} \mathbf{n} \end{math} and \begin{math} \mathbf{m} \end{math} denote tokenized sequence length of passage and question respectively. The ground truth question is denoted by \begin{math} \mathbf{Y = y_1,...,y_p} \end{math}  where \begin{math} \mathbf{p} \end{math} is the length of the ground truth question. We make use of auxiliary input i.e. answer \begin{math} \mathbf{A = a_1,...,a_b} \end{math} consisting of \begin{math} \mathbf{b} \end{math} tokens for the answer prompting mode. We make use of a sequence to sequence architecture specifically a transformer encoder and decoder network along with various methods of using answer information consisting of a custom product method involving encoder outputs and answer information, answer prompting and using an additional attention block in the decoder for answer information, and carry out experiments to look at the effect of these methods on generation of questions.

\section{Methodology}
We use a transformer based sequence to sequence architecture specifically BART-base for the task of question generation. We explain the architecture of BART below. 
\subsection{Encoder}

The encoder consists of  \begin{math} \mathbf{L} \end{math} = 6 encoder layers, where each layer contains a self-attention mechanism, layer normalization, feed forward layers and residual connections. The input to the encoder is passed through the embedding layer along with positional encodings, which convert tokens into an embedding representation \begin{math} \mathbf{E_o} \end{math}. The output from each layer in the encoder is passed on to the next layer, for all layers from \begin{math} \mathbf{L} \end{math} = 1 to \begin{math} \mathbf{L} \end{math} = \begin{math} \mathbf{N} \end{math} -1. For the last layer, the output of the encoder layer is passed to the decoder where it is used in the cross-attention mechanism. This output is \begin{math}\mathbf{d}\end{math} dimensional for embeddings \begin{math} \mathbf{E_o}  = e^{1}_o,…,e^{n}_o \end{math} for all \begin{math} \mathbf{n} \end{math} tokens in the passage. The self attention mechanism in the encoder consists of multiple heads, where each head captures a different representation. Output of these heads are concatenated and then multiplied by a weight matrix and then passed to the feedforward layer. Input to the attention block consists of queries \begin{math} \mathbf{Q \,  \varepsilon \,} \mathbb{R}^{d_q} \end{math}, keys \begin{math} \mathbf{K \,  \varepsilon \,} \mathbb{R}^{d_k} \end{math}, and values \begin{math} \mathbf{V \,  \varepsilon \,} \mathbb{R}^{d_v} \end{math} where \begin{math} \mathbf{d_q} \end{math}, \begin{math} \mathbf{d_k} \end{math}, and \begin{math} \mathbf{d_v} \end{math} are the dimensions of queries, keys and values respectively. The attention mechanism works on the principle of similarity as a dot product is carried out between keys \begin{math} \mathbf{K} \end{math} and queries \begin{math} \mathbf{Q} \end{math} and softmax is applied. For the self-attention mechanism, queries, keys and values come from source text itself. 
\begin{equation} Attention(Q, K, V) = softmax(\frac{QK^T}{\sqrt{d_k}})V \end{equation}
\subsection{Decoder}
The decoder consists of  \begin{math} \mathbf{L} \end{math} = 6 layers, each layer similar to the other, consisting of masked self-attention mechanism, masked cross attention mechanism, layer normalization and feed forward layers. Input to the decoder is  \begin{math} \mathbf{Y} \end{math} which is then passed to an embedding layer and positional encoding layer to get question embeddings  \begin{math} \mathbf{E_y} \end{math}. This output is \begin{math} \mathbf{d} \end{math} dimensional for embeddings \begin{math} \mathbf{E_Y = e_y^{1},…,e_y^{m}} \end{math} for all  tokens in the question. The attention mechanism in the decoder is similar to the encoder. The input to the attention mechanisms are masked so that the decoder does not attend to positions after the current time step. The output from each layer in the decoder is passed on the next layer, for all layers from \begin{math} \mathbf{L} \end{math} = 1 to \begin{math} \mathbf{L} \end{math} = \begin{math} \mathbf{N} \end{math} -1. For the last layer, the output is passed to a linear layer over which softmax is applied. The self attention block uses the decoder input for queries, keys and values. For the cross-attention mechanism, queries are obtained from the previous decoder layer for \begin{math} \mathbf{L} \end{math} = 2 to \begin{math} \mathbf{L} \end{math} = \begin{math} \mathbf{N} \end{math} and using the ground truth for \begin{math} \mathbf{L} \end{math} = 1, whereas the keys and values are calculated from the encoder outputs \begin{math} \mathbf{E_o} \end{math}. We carry out multiple experiments each with a different way of incorporating answer information in the architecture. During each of these experiments, the hyper-parameters for the baseline architecture were kept constant, to get consistent results and seeding was performed. We explain the different methods for incorporating answer information below. 

\subsubsection{Answer Prompting}
In this mode, we represent the answer token as \begin{math} \mathbf{A = a1,...ab} \end{math}. We concatenate the article tokens to answer tokens and truncate the final token representation to a max length of 512 tokens. Thus, \begin{math} \mathbf{X’ = A_{1:b}X_{1:n}} \end{math} This is directly fed into the encoder and the questions are generated as is from the decoder. The idea behind this is to let the self attention mechanism in the encoder calculate the similarity between the passage token and the answer tokens to aid the decoder in generating the question. We denote this mechanism as AP.

\subsubsection{Answer Attention}
We introduce an additional block, called the answer attention mechanism in the decoder. Similar to  cross-attention, the answer attention block calculates correlation between answer embedding and question token representation. We get the answer embedding \begin{math} \mathbf{E_a \,  \varepsilon \,} \mathbb{R}^{d} \end{math} by passing the answer token through the encoder and getting the mean over all the tokens to get a single representation for the answer. Thus, as per equation 1, queries \begin{math} \mathbf{Q} \end{math} are obtained from the previous layer representation, whereas the keys \begin{math} \mathbf{K} \end{math}  and values \begin{math} \mathbf{V} \end{math} are answer embeddings \begin{math} \mathbf{A} \end{math}. We denote this mechanism as AA.

\subsubsection{Answer vicinity}
Instead of providing a separate answer representation to the model, we select the sentences of the passage that contain the answer and feed it to the transformer model. We denote this mechanism as using related sentences, i.e. RS. 

\subsubsection{Using product encoder outputs and answer information}
In this mode, we align  encoder outputs \begin{math} \mathbf{E_o} \end{math} with answer embeddings \begin{math} \mathbf{A} \end{math}. We perform matrix multiplication for \begin{math} \mathbf{A} \end{math}  and \begin{math} \mathbf{E_o} \end{math}  and apply softmax over the output to get a weighted distribution of answer information over the source text. We then multiply the Encoder outputs \begin{math} \mathbf{E_o} \end{math}  with this distribution and then pass this to the cross attention block of decoder. We experiment with a multiplication constant denoted by \begin{math} \mathbf{k} \end{math}. We denote this method as custom product method i.e. CP.

\section{Experimental Setup}
We use SQUAD 1.1 dataset for our experiment on question generation. The dataset consists of 100000+ question answer pairs. Since the test set for SQUAD 1.1 is not publicly available, we use the split from \citet{zhou2018sequential} for train, test and validation. We tokenize the data using BART tokenizer from huggingface. We set the maximum length of the paragraph to 512 tokens, each question to 128 tokens, and each answer to 32 tokens. We use the same hyperparameters for all the different answer information methods in order to ensure consistency across the experiments. Moreover, we use the same seed for all the experiments across all libraries. For the model, we use BART-base as our transformer architecture. BART-base was chosen over BART-large due to insufficient compute resources for training the model. BART-base consists of 6 layers of encoder and decoder respectively. The hidden dimension \begin{math} \mathbf{d} \end{math} is 768 dimensional for BART-base. A vocabulary size \begin{math} \mathbf{V} \end{math} of 50265 is used. The baseline architecture consists of 139M parameters. We use cross entropy loss for our training objective. A batch size of 8 is used for training and validation. We use a learning rate of 1e-05 with a linear learning rate scheduler. We train the model for a total of 100k steps on a NVIDIA Quadro 16GB GPU where each experiment took roughly 30 hours. A beam size of 4 was used in beam search for decoding.

\section{Results and Analysis}
We use ROUGE-L and METEOR score as metrics for evaluating the generated questions. ROUGE-L measures the longest common subsequence between the ground truth questions and the generated questions. METEOR score works on the principle of harmonic mean, where a harmonic mean of precision and recall is carried out with recall having a higher weight than precision. Although METEOR is commonly used for machine translation, it can also be used for question generation. Additionally, to verify if the question generated is according to the answer provided, we employ a question answering model denoted by QA. \begin{equation} A' = QA(X,Y') \end{equation} We compare A and A’ for each of the models, to check what percentage of generated questions when answered, have the same answer as that in the dataset. The percentage of questions having the exact same answer for each of the models is shown in Table \hyperref[tab:table 1]{ \ref*{tab:table 1}}. 

We carry out experiments with various answer information strategies as described above, as well as their combinations as shown in Table \hyperref[tab:table 1]{ \ref*{tab:table 1}}. We observe that questions generated from AP gave the best results for the chosen metrics. Adding RS to AP led in reduction across all metrics over \begin{math} BART_{AP} \end{math}. However, adding RS to CP led to an improvement over \begin{math} BART_{CP} \end{math}. Adding RS to the CP + AP setting slightly improves ROUGE-L but shows a mild reduction across the other metrics. We also experimented with the multiplication constant, and noticed that a value of 1e2 gave us the best results for this setting. 

\begin{table}
\begin{adjustbox}{width=\columnwidth,center}
\begin{tabular}{llll}
\hline
\textbf{Model} & \textbf{ROUGE-L} & \textbf{METEOR} & \textbf{Answering} \\
& & &\textbf{Accuracy (\%)} \\
\hline
BART$_{AA}$ & 0.3128 & 0.3019 & 17.11 \\
BART$_{CP}$  & 0.3701 & 0.3618 & 30.9 \\ 
BART$_{AP}$ & \textbf{0.4565} & \textbf{0.4595} & \textbf{57.33}\\
BART$_{CP + RS}$ & 0.3939 & 0.3893 & 34.61\\
BART$_{AP + RS}$ & 0.4506 & 0.4513 & 56.53\\
BART$_{CP + AP}$ & 0.4479 & 0.4488 & 54.42\\
BART$_{AP + RS + CP}$ & 0.4489 & 0.4453 & 54.07\\ 
\hline
\end{tabular}
\end{adjustbox}
\caption{Metric values for SQUAD 1.1 dataset}
\label{tab:table 1}
\end{table}

\section{Conclusion and Future Work}
We carry out experiments with multiple answer information methods like answer prompting, using a custom product method between answer embeddings and encoder outputs, choosing sentences from the input paragraph that have answer related information, and using a separate cross-attention attention block in the decoder which attends to the answer. Due to limited compute, we were unable to work with different transformer architectures and explore which answer information method works best. It would be interesting to observe a correlation between the base architecture and performance on of a certain answer information method on a given metric. 

\nocite{Ando2005,borschinger-johnson-2011-particle,andrew2007scalable,rasooli-tetrault-2015,goodman-etal-2016-noise,harper-2014-learning}

\bibliography{anthology,custom}
\bibliographystyle{acl_natbib}

\end{document}